%% file: root.tex
\documentclass[letterpaper, 10 pt, conference]{ieeeconf}  

\IEEEoverridecommandlockouts                              

\usepackage{booktabs}
\usepackage{multirow}
\usepackage{amssymb}
\usepackage{graphicx}
\usepackage{amsmath}
\usepackage[table]{xcolor}
\usepackage{url}
\providecommand{\CircleArrowright}{\circlearrowright}

\newcommand{\methodname}{JOP-VLN}
\definecolor{LightGray}{gray}{0.97}
\definecolor{LightBlue}{rgb}{0.97,0.985,1.0}

\definecolor{arrowlightgreen}{HTML}{97D077}  
\definecolor{arroworange}{HTML}{FFB570}      
\definecolor{arrowbrightgreen}{HTML}{00FF00} 

\title{\LARGE \bf
\methodname: Joint On-and-Off Policy Learning for \\ Vision-and-Language Navigation
}

\author{Qingrong He$^{1}$, Lin Zhao$^{1}$, Kevin Zheng$^{1}$, and Liang Lin$^{1\dagger}$%
\thanks{$^{1}$All authors are with Joy Future Academy.}
\thanks{$^{\dagger}$Corresponding author.}%
}

\begin{document}

\maketitle
\thispagestyle{empty}
\pagestyle{empty}


\input{sec/0_abstract}
\input{sec/1_intro}
\input{sec/2_related_work}
\input{sec/3_method}
\input{sec/4_experiments}
\input{sec/5_conclusion}

\addtolength{\textheight}{-2cm}   


\bibliographystyle{ieeetr}
\bibliography{refs}

\end{document}

%% file: sec/0_abstract.tex
\begin{abstract}

Vision-and-Language Navigation (VLN) necessitates an embodied agent to navigate in the physical world by adhering to natural language instructions. Recent advancements in Vision-Language Models (VLM) have propelled the development of VLM-based VLN methods with two predominant paradigms: (1) imitation learning (IL) on expert demonstrations, followed by the Dataset Aggregation (DAgger) algorithm to bolster error recovery capabilities; (2) reinforcement learning (RL) driven by verifiable rewards to enhance reasoning and exploration. A notable gap is the absence of integration between these two distinct paradigms. This paper introduces JOP-VLN, a novel VLN framework that synergistically combines off-policy imitation learning and on-policy exploration within a three-stage training pipeline. Initially, IL is employed on expert demonstrations to acquire basic navigation skills. Subsequently, the DAgger algorithm is utilized to generate heuristic exploration trajectories, which are then used for imitation learning to improve error recovery capabilities. Finally, a joint on-and-off policy learning framework is implemented, featuring high-entropy trajectory sampling to enhance RL training efficiency and an error-correction-prioritized trajectory sorting strategy for effective error correction. Extensive experiments demonstrate the efficacy of JOP-VLN, achieving success rates of 69.9\% and 68.0\% on the VLN-CE R2R and RxR benchmarks, respectively, setting a new state-of-the-art on R2R.
Project page: \url{https://qingrongh.github.io/JOP-VLN}.

\end{abstract}

%% file: sec/1_intro.tex
\section{Introduction}
\label{sec:intro}
Vision-and-Language Navigation (VLN) represents a fundamental capability for developing autonomous embodied agents, requiring them navigate physical environments by following natural language instructions. 
This task poses significant challenges in modeling cross-modal alignment between textual and visual cues, processing temporal visual history, and inferring future actions.
To address these issues, recent studies \cite{zheng2024navillm, cheng2024navila, zhang2024navid, zhang2024uninavid, wei2025streamvln} have adapted Vision-Language Models (VLM) into navigation decision-making, leveraging their rich inherent priors in language understanding and visual perception.

Prior research \cite{cheng2024navila, zhang2024navid, zhang2024uninavid, wei2025streamvln} typically leverages imitation learning to train agents on navigation data from diverse sources.
Despite substantial advancements, this paradigm suffers from covariate shift, where cumulative errors arise as the agent encounters out-of-distribution states not seen during training, thereby incurring poor generalization. 
To mitigate this, the Dataset Aggregation (DAgger) algorithm~\cite{ross2011dagger} has been applied as an effective IL strategy to augment the training set with oracle-labeled off-policy trajectories. 
Nevertheless, they remain constrained by a heavy reliance on expert demonstrations, limiting their generalization to unseen environments.

Inspired by the success of Reinforcement Learning from Verifiable Rewards (RLVR)~\cite{deepseek-math, yu2025dapo, gspo, sapo}, several studies focus on applying reinforcement techniques, such as GRPO \cite{deepseek-math}, following the IL phase. 
For instance, VLN-R1 \cite{qi2025vln-r1} and OctoNav \cite{gao2025octonav} employ GRPO training by leveraging existing trajectories. 
Although effective in improving action prediction, the use of in-distribution data fails to adequately address covariate shift during inference. 
This naturally motivates us to integrate the strengths of on-policy GRPO and off-policy DAgger training.

To this end, we introduce JOP-VLN, a novel VLN model that leverages a joint on-and-off policy learning paradigm via three progressive training stages.
In the initial stage, the model is trained on diverse tasks, including navigation data and trajectory summarization, to acquire fundamental navigation skills. 
Subsequently, the model is fine-tuned using off-policy trajectories collected by the learned policy via the DAgger algorithm to improve its error-recovery capabilities.
To further enhance generalization, the final stage integrates off-policy imitation learning with on-policy reinforcement learning, drawing inspiration from CHORD \cite{MIXCHORD}.
This involves jointly optimizing the on-policy GRPO objective and the IL objective on off-policy trajectories collected via the DAgger algorithm.
Our method distinguishes itself from CHORD \cite{MIXCHORD} through several VLN-specific techniques.
First, we introduce a trajectory sampling strategy to exclude low-entropy trajectories, preventing the model from becoming overconfident and degrading model performance.
Second, a trajectory sorting strategy is employed to prioritize error-prone samples, allowing the model to focus on its frequent failures and facilitating a seamless transition from IL to RL.

Extensive experiments demonstrate the superiority of JOP-VLN, which achieves a new state-of-the-art on the Val-Unseen split of R2R~\cite{anderson2018r2r} at 69.9\% SR.
It also obtains 68.0\% SR and 59.3\% SPL on RxR~\cite{ku2020rxr} Val-Unseen split, demonstrating its excellent long-horizon decision-making capability.
Additionally, we conduct real-world experiments in challenging indoor and outdoor scenarios under a few-shot adaptation setting, and the results validate the effectiveness of our proposed method in real-world deployment.

In summary, our contributions are as follows:
\begin{itemize}
\item We introduce JOP-VLN, a novel VLN framework that leverages a joint on-and-off policy learning paradigm via three progressive stages.
\item We devise two techniques to adapt the CHORD framework to the VLN setting, \textit{i.e.}, a trajectory sampling strategy that retains high-entropy trajectories and a trajectory sorting strategy that prioritizes error-prone samples.
\item We conduct extensive experiments in both simulation and real-world environments. Our method achieves a new state-of-the-art of 69.9\% SR on the R2R Val-Unseen split,  demonstrating the superiority of our training framework.
\end{itemize}

%% file: sec/2_related_work.tex
\section{Related Work}
\label{sec:Related Work}

\subsection{Vision-and-Language Navigation (VLN)}
As a fundamental task in embodied AI, VLN  \cite{anderson2018vision, krantz_vlnce_2020, ku2020room, qi2020reverie, thomason2020vision} has spurred extensive interest into various learning techniques, ranging from multi-task learning \cite{zheng2024navillm, cheng2024navila, zhang2024uninavid, wei2025streamvln} and curriculum learning \cite{zhu2020babywalk, zhang2021curriculum} to reinforcement learning \cite{qi2025vln-r1, krantz_vlnce_2020, he2021landmark, wang2019reinforced, zhang2025activevln, huang2025mobilevla}.
Our work is closely related to studies of reinforcement learning for VLN.
Early efforts focus on designing heuristic rewards, including arrival signal \cite{krantz_vlnce_2020}, instruction-landmark alignment \cite{he2021landmark}, and navigation progress \cite{wang2019reinforced}.
More recently, there has been an incresing emphasis on leveraging RL to optimize Vision-Language Models (VLMs) for VLN.
Specifically, VLN-R1 \cite{qi2025vln-r1} and OctoNav~\cite{gao2025octonav} incorporates Group Relative Policy Optimization (GRPO) by utilizing a matching score between the ground truth and prediction as the reward signal to bolster navigation performance.
ActiveVLN \cite{zhang2025activevln} extends the GRPO process to multi-turn active exploration.
In contrast with these work, JOP-VLN structurally fuses on-policy RL exploration with off-policy IL supervision. This allows the agent to learn simultaneously from expert demonstrations and environmental exploration signals.

\subsection{Navigation with Vision-Language Models}
Motivated by the remarkable success of Vision-Language Models (VLMs), recent research \cite{zheng2024navillm, cheng2024navila, zhang2024navid, zhang2024uninavid, wei2025streamvln, long2024instructnav, wang2025dynam3d, zhou2024navgpt} has adapted VLMs to embodied navigation for improved generalization to unseen scenarios.
Pioneering works such as NaVid \cite{zhang2024navid} and Uni-NaVid \cite{zhang2024uninavid} leverage multi-task learning across diverse data sources to enhance VLM-based navigation models.
Subsequently, NaVILA \cite{cheng2024navila} introduces a heuristic pipeline to automatically annotate human videos, facilitating effective sim-to-real transfer.
Furthermore, StreamVLN \cite{wei2025streamvln} and CorrectNav \cite{yu2025correctnav} leverage the pre-collected trajectories for enhancing the error-recovery capabilities during the navigation process.
Despite these advancements, imitation learning from expert demonstrations may restrict generalization capabilities to mere memorization, thus limiting robustness. Incorporating on-policy RL within the off-policy IL process aids JOP-VLN in enhancing exploration while mitigating the risk of overfitting on off-policy expert data.

%% file: sec/3_method.tex
\section{Framework}
\subsection{Task Formulation}
The Vision-and-Language Navigation (VLN) task requires an agent to navigate to a target location based on a language instruction $I$.  
Specifically, at each step $t$, the agent first perceives the current visual observation $o_t$, and then predicts the next action chunk  $a_t$ based on the entire observation and action histroy $\{o_{1:t},a_{1:t-1}\}$, denoted by $a_t \sim \pi_{\theta}(\cdot \vert I, \{o_{1:t},a_{1:t-1}\})$.
Following previous work \cite{wei2025streamvln}, each chunk consists of four consecutive actions within a discrete space encompassing \texttt{forward}, \texttt{left}, \texttt{right}, and \texttt{stop}.

\subsection{Model Architecture}
\label{sec:multi-turn}

\begin{figure}[!h]
    \centering
    \includegraphics[width=\columnwidth]{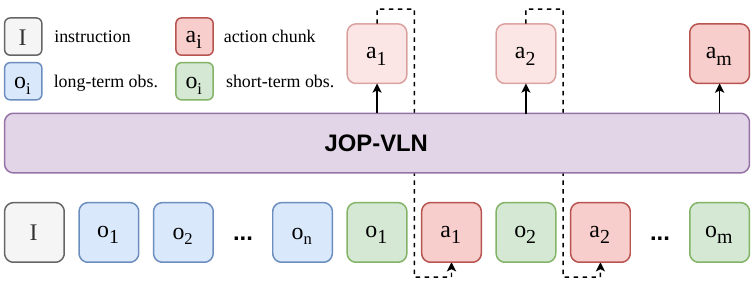}
    \caption{Multi-turn Dialogue Paradigm.}
    \label{fig:multi-turn}
\end{figure}

Our model is built upon a large vision-language model, \textit{i.e.}, Qwen3-VL~\cite{Qwen3-VL}.
As shown in Figure \ref{fig:multi-turn},
our model adopts a multi-turn dialogue paradigm following previous work \cite{wei2025streamvln}, where the model processes the language instruction $I$, a long-term memory of $n$ frames, and a short-term memory of $m$ frames.
Specifically, at step $t$, the long-term memory consists of frames uniformly sampled from the distant history $\{o_{1:t-k-1}\}$, and the short-term memory maintains the most recent interleaved sequences $\{o_{t-k:t}, a_{t-k:t-1}\}$, where $k=t\%m$.
Upon generating an action chunk, the model receives a new observation and appends both to the current sequence. 
This design facilitates efficient KV cache reuse, enabling the sequential prediction of action chunks within a near-linear inference budget. 
Upon the short-term buffer reaching its capacity $m$, the long-term memory is updated via uniform resampling, while the short-term memory is flushed and reset to the current observation.

\section{On-and-Off Policy Training}
\label{sec:training}
We devise an on-and-off policy training paradigm consisting of three progressive stages.
In the initial stage, the model is trained in a multi-task manner to acquire the fundamental navigation skills.
Then, we generate DAgger trajectories using a heuristic strategy, which are subsequently used for imitation training, thereby enhancing its error-recovery capabilities.
In the final stage, we perform on-and-off policy training following the CHORD \cite{MIXCHORD} framework.
This involves on-policy GRPO training and off-policy imitation learning on the DAgger trajectories.

\subsection{Stage 1. Multi-task Training}
\begin{figure}[t]
    \centering
    \vspace{+5pt}
    \includegraphics[width=0.9\columnwidth]{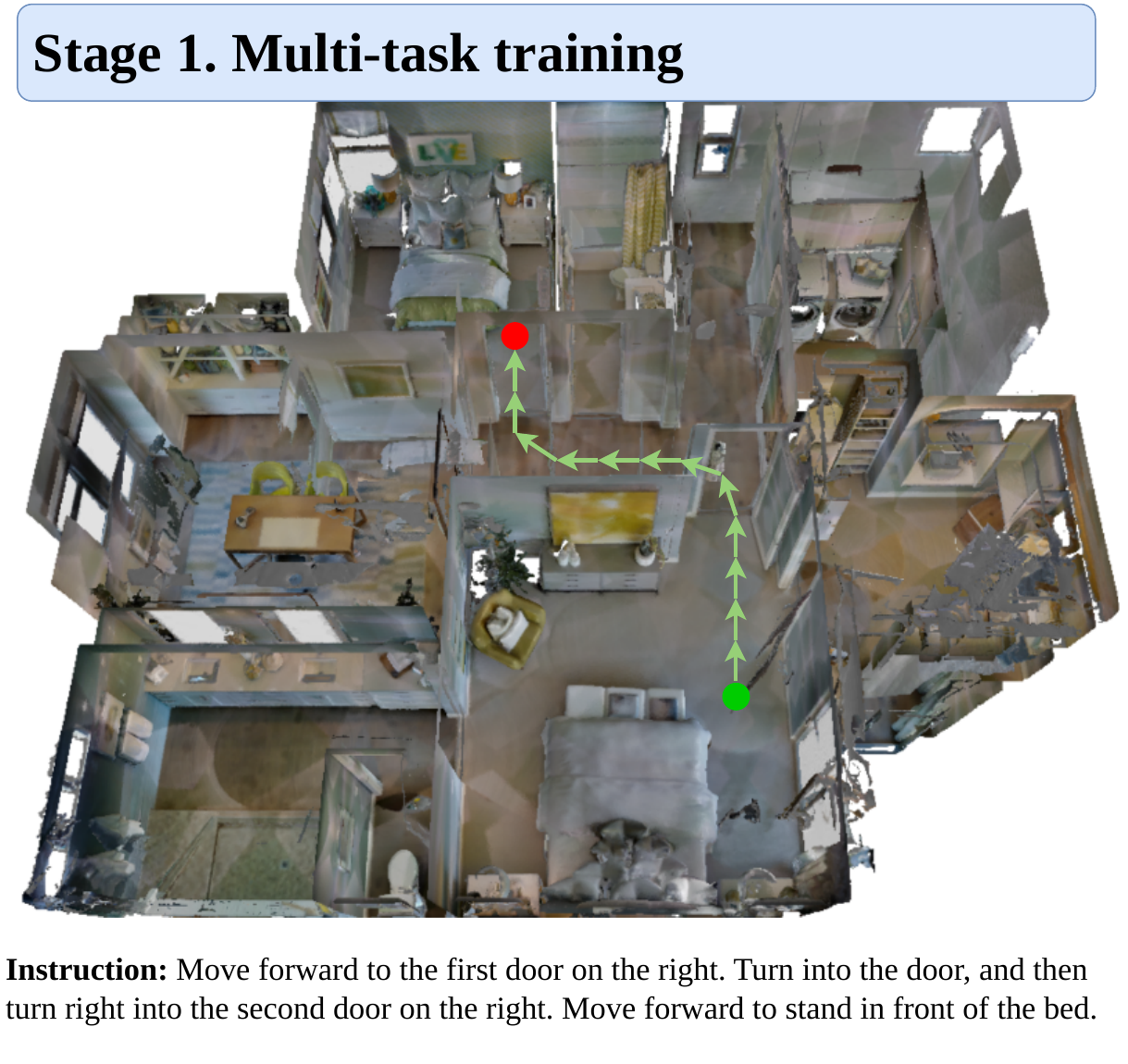}
    \vspace{-15pt}
    \caption{Stage 1. Multi-task training.}
    \label{fig:stage1}
\end{figure}

As shown in Figure~\ref{fig:stage1}, in this stage, we co-train our model on the tasks of action prediction and trajectory summarization.
Specifically, for action prediction, the model is tasked with predicting action chunks within the multi-turn dialogue paradigm detailed in Section~\ref{sec:multi-turn}. 
For trajectory summarization, the model is required to synthesize textual instructions conditioned on sampled frames from navigation videos.

\subsection{Stage 2. Imitation Learning with DAgger Trajectories.}
\begin{figure}[t]
    \centering
    \vspace{-14pt}
    \includegraphics[width=0.9\columnwidth]{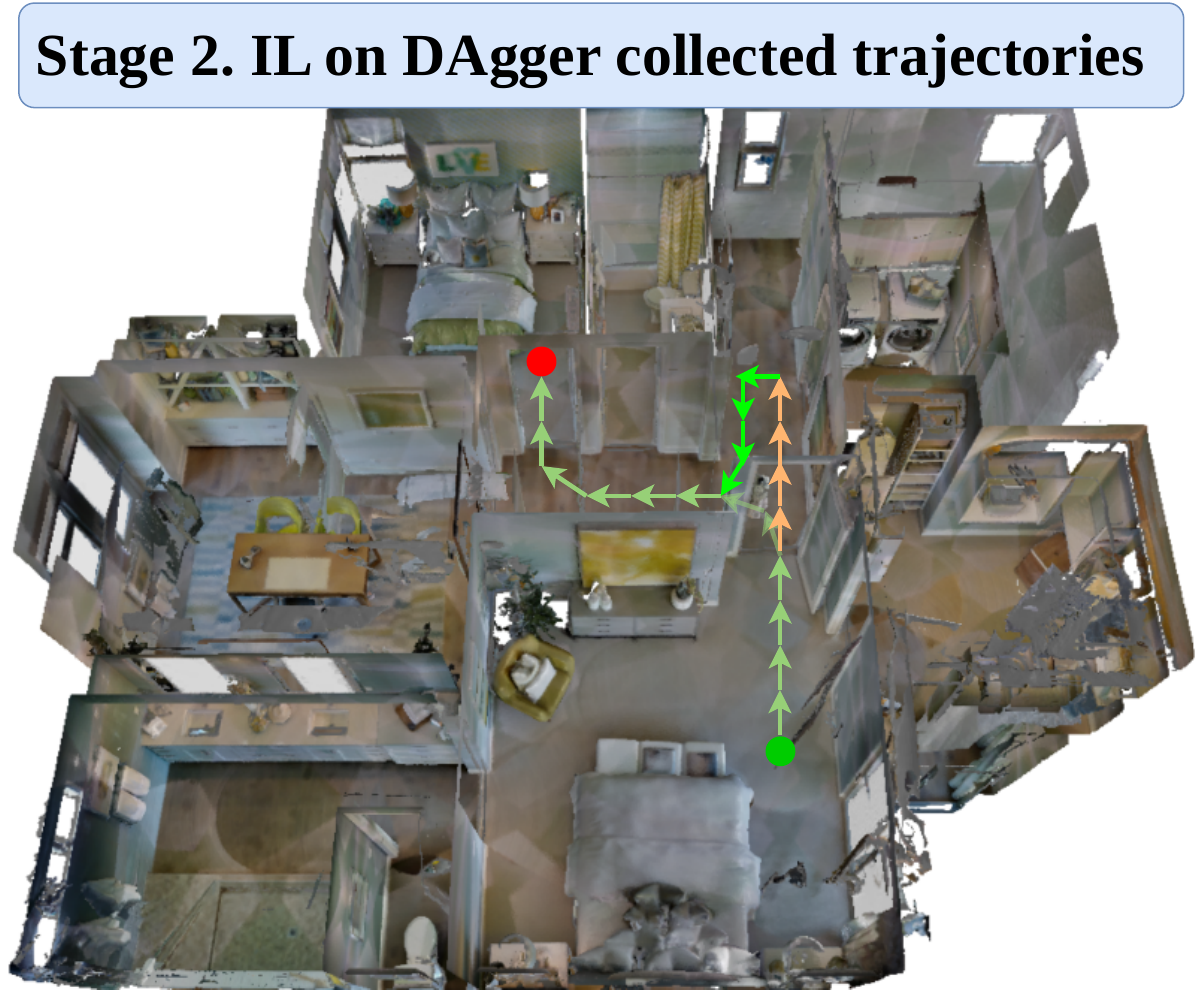}
    \vspace{-10pt}
    \caption{Stage 2. Imitation learning with DAgger trajectories.
    \textcolor{arrowlightgreen}{$\boldsymbol{\uparrow}$}denotes model-predicted actions,
    \textcolor{arroworange}{$\boldsymbol{\uparrow}$}denotes actions deviating from the optimal path, and
    \textcolor{arrowbrightgreen}{$\boldsymbol{\uparrow}$}denotes oracle-corrected actions.}
    \label{fig:stage2}
\end{figure}

The Dataset Aggregation (DAgger) algorithm \cite{ross2011dagger} is widely adopted for generating trajectories through a mixed policy that combines learned and oracle strategies. 
The implementation of StreamVLN \cite{wei2025streamvln} adapts the original framework by introducing a heuristic strategy. Specifically, this strategy prioritizes the learned policy and only triggers oracle intervention when the agent's deviation from the ground-truth trajectory exceeds a predefined threshold. 
StreamVLN's action labels may contain erroneous actions generated by the model, which can hinder the fine-tuning process. 
To mitigate this issue, following \cite{zheng2025efficient}, we replay all visited states to obtain their corresponding oracle action chunks, which are subsequently utilized for imitation learning as shown in Figure~\ref{fig:stage2}.

\subsection{Stage 3. Joint On-and-Off Policy Learning with High-entropy and Error-correction-prioritised Trajectories}
\begin{figure}[t]
    \centering
    \vspace{+5pt}
    \includegraphics[width=0.9\columnwidth]{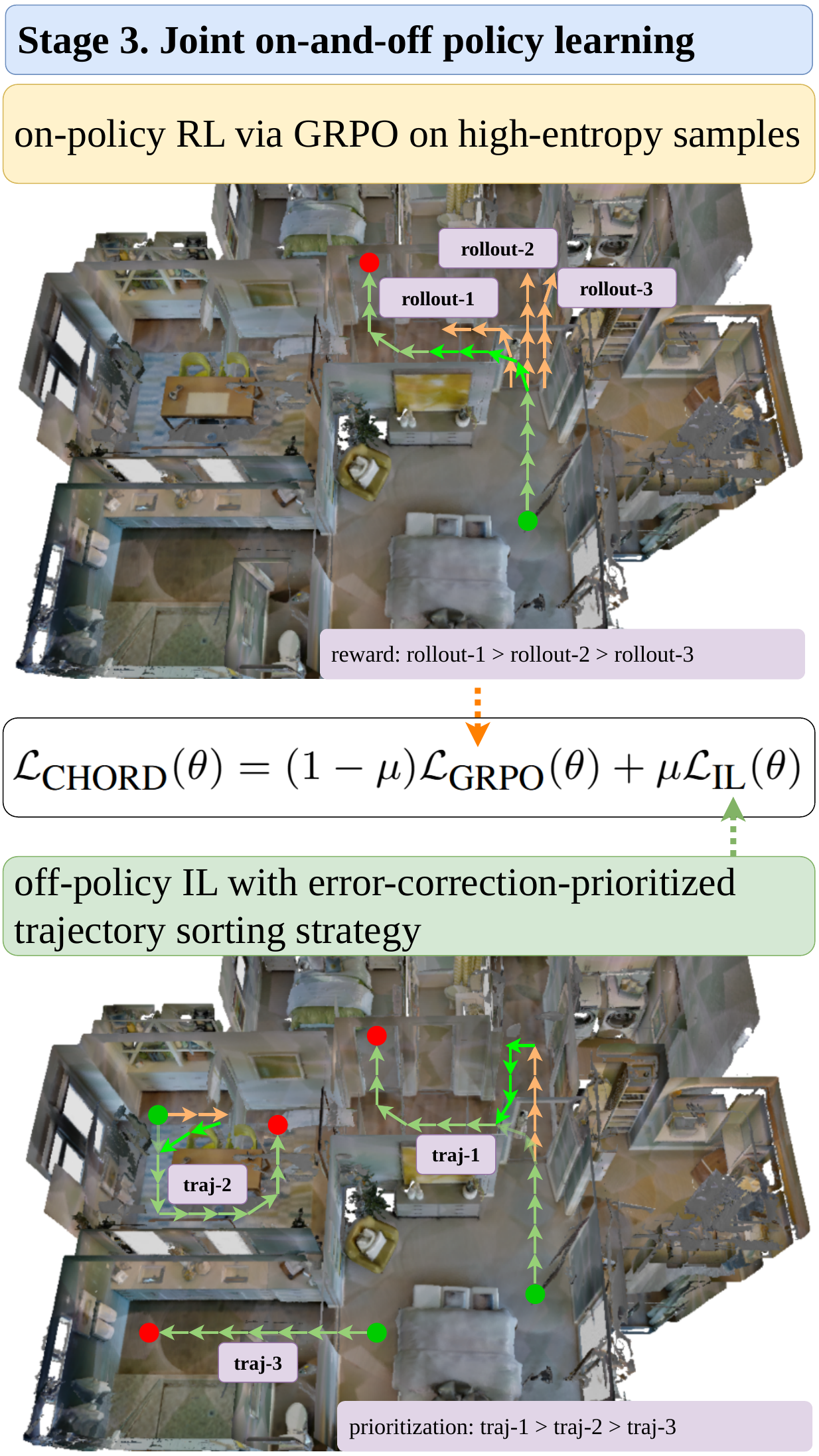}
    \caption{Stage 3. Joint on-and-off policy learning with high-entropy and error-correction-prioritised trajectories.}
    \vspace{-10pt}
    \label{fig:stage3}
\end{figure}

The DAgger algorithm is reapplied, during which the entropies of the model-predicted tokens are recorded. The CHORD~\cite{MIXCHORD} framework is employed on these DAgger collected trajectories with high-entropy trajectory sampling for GRPO and error-correction-prioritised trajectory sorting for imitation learning, as shown in Fiture~\ref{fig:stage3}.

\subsubsection{\textbf{On-policy exploration with GRPO on high-entropy trajectories}} \label{subsec:rl}
Given the input comprising language instructions, long-term and short-term memory, the policy $\pi_{\theta}$ generates $G$ responses $\{a_1^*, a_2^*, \ldots, a_G^*\}$ for each sample. The reward for each response is calculated using the following four reward functions.

\textbf{Format reward} $R_{format}$ checks whether the response adheres to the specified format, \textit{i.e.}, containing at most four consecutive discrete actions.
\begin{equation}
\footnotesize
    R_{format} =
    \begin{cases}
        1, & \text{if the response adheres to the specified format} \\
        0, & \text{otherwise}
    \end{cases}
\end{equation}

\textbf{Exact match reward} $R_{em}$ evaluates whether the response $a^*$ precisely matches the oracle action chunk $a$.
\begin{equation}
\footnotesize
    R_{em} =
    \begin{cases}
        1, & \text{if } a^* = a \\
        0, & \text{otherwise}
    \end{cases}
\end{equation}

\textbf{Position reward} $R_{pos}$ assesses the cosine similarity between $pos^*$ and $pos$, where $pos^* = (\triangle x^*, \triangle y^*, \triangle yaw^*)$ and $pos = (\triangle x, \triangle y, \triangle yaw)$ denote the change in the robot's pose after executing the predicted action chunk $a^*$ and the oracle action chunk $a$, respectively.
\begin{equation}
\footnotesize
R_{pos} =
\frac{pos^\top pos^*}
{\lVert pos \rVert_2 \, \lVert pos^* \rVert_2}
\end{equation}

\textbf{Distance reward} $R_{dis}$ quantifies the Euclidean distance between the predicted displacement $(\triangle x^*, \triangle y^*)$ and the oracle displacement $(\triangle x, \triangle y)$. This distance is then normalized by dividing it by the maximum possible distance, calculated as $2 \times c \times s$, where $c$ is action chunk size and $s$ is the forward step size.
\begin{equation}
\footnotesize
R_{dis} =
\frac{\sqrt{(\triangle x^* - \triangle x)^2 + (\triangle y^* - \triangle y)^2}}{2 \times c \times s}
\end{equation}

The advantages $\hat{A}_i$ of rewards $r = \{r_1, r_2, \ldots, r_G\}$ for the $G$ responses sampled from the current policy $\pi_{\theta}$ is calculated as:
\begin{equation}
\footnotesize
\hat{A}_i = \frac{r_i - \text{mean}(\{r_i\}_{i=1}^{G})}{\text{std}(\{r_i\}_{i=1}^{G})}
\end{equation}

The policy is optimized via the GRPO objective:
\begin{equation}
\footnotesize
\resizebox{0.9\hsize}{!}{$
\begin{split}
& L_{\mathrm{GRPO}}(\theta) = 
\mathbb{E}_{i=1}^G \Biggl[
    \min \Biggl(
        \frac{\pi_\theta(a_i^* \mid I)}{\pi_{\theta_{\mathrm{old}}}(a_i^* \mid I)} \hat{A}_i, \\
        & \operatorname{clip}\!\left(
            \frac{\pi_\theta(a_i^* \mid I)}{\pi_{\theta_{\mathrm{old}}}(a_i^* \mid I)},
            1 - \epsilon,\; 1 + \epsilon
        \right) \hat{A}_i
    \Biggr) 
    - \beta \cdot D_{\mathrm{KL}}\!\left(\pi_\theta \,\|\, \pi_{\text{ref}}\right)
\Biggr]
\end{split}
$}
\label{eq:grpo_loss}
\end{equation}

\noindent where $I$ is the input to the model, $a_i^*$ is the $i$-th response, and $\pi_{\text{ref}}$ is the reference model. In our implementation, the token-level policy gradient loss is applied for enhancing training stability~\cite{yu2025dapo}.

\textbf{High-entropy trajectory sampling.} Previous studies \cite{wang2025beyond, le2026rlzvp} have identified the critical role of high-entropy minority tokens in reinforcement learning training. Remarkably, training with only 20\% of these high-entropy tokens can achieve performance that is comparable to or even surpasses that obtained with 100\% of the tokens. Inspired by these findings, we adopt a high-entropy trajectory sampling strategy. During the DAgger trajectory collection process, we record the entropy of each model-predicted action token and conduct GRPO training using only the top 20\% of samples with the highest entropy values. Employing the high-entropy trajectory sampling strategy fosters exploration and prevents the model from entering a ``zero-gradient" state caused by low-entropy samples, where the advantage collapses to zero, eliminating training signals and rendering rollouts uninformative.

\subsubsection{\textbf{Off-policy imitation learning}}
\label{subsec:il}

Let $\mathcal D=\{(I,a)\}$ denote a corpus of expert demonstrations collected using the DAgger algorithm, where $a$ denotes the oracle actions corresponding to the input $I$, organized within the multi-turn dialogue paradigm as described in Section~\ref{sec:multi-turn}. Imitation learning aims to minimize the sentence-level cross-entropy:

\begin{equation}
\footnotesize
  \mathcal{L}_{\mathrm{IL}}(\theta)
  \;=\;
  \mathbb{E}_{(I, a)\sim\mathcal D}
  \bigl[-\log \pi_\theta\bigl(a \mid I\bigr)\bigr].
\label{eq:il_loss}
\end{equation}

\subsubsection{\textbf{Joint on-and-off policy learning with CHORD framework prioritizing error-correction trajectories}}
\label{subsec:chord}
Following CHORD~\cite{MIXCHORD}, imitation learning is integrated as a dynamically weighted auxiliary objective within the on-policy reinforcement learning process. This is achieved through a combined loss function that minimizes a weighted sum of both RL and IL losses:

\begin{equation}
\footnotesize
\mathcal{L}_{\text{CHORD}}(\theta) = (1-\mu) \mathcal{L}_{\text{GRPO}}(\theta) + \mu \mathcal{L}_{\text{IL}}(\theta)
\label{eq:chord_loss}
\end{equation}

\noindent where $\mathcal{L}_{\text{GRPO}}(\theta)$ is the GRPO loss defined in~\eqref{eq:grpo_loss}, $\mathcal{L}_{\text{IL}}(\theta)$ denotes the IL loss defined in~\eqref{eq:il_loss}. The hyperparameter $\mu \in [0, 1]$ regulates the trade-off between imitation learning and reinforcement learning.

\begin{figure}[!h]
    \centering
    \includegraphics[width=\columnwidth]{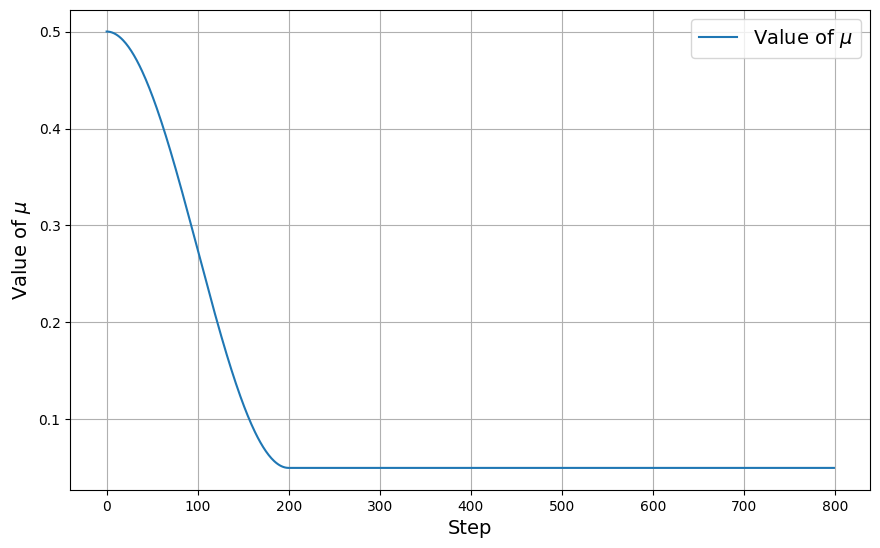}
    \vspace{-20pt}
    \caption{$\mu$ schedule with cosine decay.}
    \label{fig:mu_schedule}
\end{figure}

A decay schedule for $\mu$ is implemented to facilitate a seamless transition from a balanced imitation-reinforcement combination to on-policy dominated optimization, as illustrated in Figure~\ref{fig:mu_schedule}. The training process initiates with $\mu = 0.5$, assigning equal weight to off-policy imitation and on-policy exploration so that the model can benefit from expert demonstrations while stabilizing the early RL updates. As training advances, $\mu$ is progressively reduced to $0.05$, thereby shifting the emphasis towards on-policy exploration while reducing the risk of overfitting on off-policy expert data.

\textbf{Error-correction-prioritised trajectory sorting.}
The dynamic scheduling strategy for $\mu$ prioritizes imitation learning during the early stage of training. To improve the model's error recovery capabilities, we devise an error-correction-prioritised trajectory sorting strategy. 
Specifically, for a multi-turn dialogue trajectory sample as described in Section~\ref{sec:multi-turn}, we identify which actions within the trajectory are corrected by the oracle while the model made mistakes. The learning value of a trajectory is determined by the proportion of these corrected action tokens. These expert demonstrations are then sorted in descending order of learning value for the model to perform imitation learning. This ensures that the model more frequently learns from trajectories where it previously made mistakes during the DAgger trajectory collection process.

%% file: sec/4_experiments.tex
\section{Experiment}
\subsection{Experimental Setup}

\textbf{Simulation benchmarks and evaluation metrics.}
We evaluate JOP-VLN on two widely used VLN-CE~\cite{krantz_vlnce_2020} benchmarks, including R2R~\cite{anderson2018r2r} and RxR~\cite{ku2020rxr}. 
Both datasets are built upon Matterport3D~\cite{chang2017matterport3d} scenes within the Habitat simulator~\cite{szot2021habitat}.
Following prior work \cite{cheng2024navila,zhang2024navid, zhang2024uninavid}, we report standard VLN metrics, including navigation error (NE), success rate (SR), oracle success rate (OS), success weighted by path length (SPL), and normalized dynamic time warping (nDTW).
All evaluations are conducted on the Val-Unseen split where the scenes are not seen during training.

\textbf{Real-World Experiment Configuration.} 
Our real-world experiments are conducted on a Unitree Go2 quadruped robot, equipped with an Intel RealSense D435i RGB-D camera. 
\methodname{} operates on a remote workstation utilizing an L40S GPU. 
During the navigation process, the robot transmits the captured RGB frames to the server over a WiFi connection. 
Subsequently, the server processes the received data and sends the predicted action commands back to the robot for execution. 
The system achieves an average inference time of 0.3s per action chunk, with a communication delay of 1.2s. Since each chunk contains up to 4 sequential discrete actions, the per-action execution rate reaches up to 2.67Hz. The communication delay dominates the end-to-end latency and can be further reduced through on-device deployment.

\subsection{Implementation Details}
\textbf{Training Data.}
During the first stage, we aggregate 793.4k samples for action prediction from R2R \cite{anderson2018r2r}, RxR \cite{ku2020rxr}, EnvDrop \cite{tan2019envdrop}, and ScaleVLN's subset \cite{wang2023scalevln}, and 30.8k video-instruction pairs from R2R and RxR for trajectory summarization.
In the second stage, we synthesize 11.3k trajectories based on the R2R and RxR annotations.
By combining them with the trajectories from the R2R, RxR, and ScaleVLN's subset, we obtain 453.3k training samples.
In the last stage, we collect an additional 7.4k DAgger trajectories. 
From this, 33.3k high-entropy samples are employed for GRPO training, while 30.0k samples are used for imitation learning.

\textbf{Training Details.}
JOP-VLN is built upon the Qwen3-VL-8B-Instruct~\cite{Qwen3-VL} backbone and optimized using AdamW.
During the first two stages, the peak learning rates for the language model, projector, and vision encoder are set to 2e-5, 1e-5, and 5e-6, respectively. 
In the third stage, these rates are adjusted to 1e-6, 5e-7, and 5e-8, respectively. 
For the GRPO objective, we generate eight rollouts per sample, with the KL coefficient $\beta$ and clipping parameter $\epsilon$ configured at $0.1$ and $0.2$, respectively.
The parameter $\mu$ follows a dynamic schedule, decaying from 0.5 to 0.05 over the first 200 training steps and remaining constant thereafter.

\begin{table*}[]
\centering
\small
\vspace{+15pt}
\caption{
Comparison with state-of-the-art methods on VLN-CE R2R and RxR Val-Unseen split.
The ``Observation Encoder'' inputs include panoramic (Pano.), odometry (Odo.), depth image (Depth), and single RGB image (S.RGB). $*$ denotes methods utilizing the waypoint predictor from~\cite{hong2022bridging}.
}
\resizebox{1.\linewidth}{!}{
\setlength{\tabcolsep}{3mm}
\begin{tabular}{l|cccc|cccc|cccc}
\toprule
\multirow{2}{*}{Method} & \multicolumn{4}{c|}{Observation Encoder} & \multicolumn{4}{c|}{R2R Val-Unseen} & \multicolumn{4}{c}{RxR Val-Unseen} \\ 
\cmidrule(lr){2-5} \cmidrule(lr){6-9} \cmidrule(lr){10-13}
      & Pano. & Odo. & Depth & S.RGB & NE$\downarrow$ & OS$\uparrow$ & SR$\uparrow$ & SPL$\uparrow$ & NE$\downarrow$ & SR$\uparrow$ & SPL$\uparrow$ & nDTW$\uparrow$ \\ 
\midrule
HPN+DN$^*$~\cite{krantz2021waypoint} & $\checkmark$ & $\checkmark$ & $\checkmark$ &  & 6.31  & 40.0  & 36.0  & 34.0  & - & - & - & - \\
CMA$^*$~\cite{hong2022bridging}      & $\checkmark$ & $\checkmark$ & $\checkmark$ &  & 6.20  & 52.0  & 41.0  & 36.0  & 8.76 & 26.5 & 22.1 & 47.0 \\
VLN$\protect\CircleArrowright$BERT$^*$~\cite{hong2022bridging} & $\checkmark$ & $\checkmark$ & $\checkmark$ &  & 5.74  & 53.0  & 44.0  & 39.0  & 8.98 & 27.0 & 22.6 & 46.7 \\
Sim2Sim$^*$~\cite{krantz2022sim}    & $\checkmark$ & $\checkmark$ & $\checkmark$ &  & 6.07  & 52.0  & 43.0  & 36.0  & - & - & - & - \\
GridMM$^*$~\cite{wang2023gridmm}    & $\checkmark$ & $\checkmark$ & $\checkmark$ &  & 5.11  & 61.0  & 49.0  & 41.0  & - & - & - & - \\
ETPNav$^*$~\cite{an2023etpnav}      & $\checkmark$ & $\checkmark$ & $\checkmark$ &  & 4.71  & 65.0  & 57.0  & 49.0  & 5.64 & 54.7 & 44.8 & 61.9 \\ 
ScaleVLN$^{*}$~\cite{wang2023scalevln} & $\checkmark$ & $\checkmark$ & $\checkmark$ &  & 4.80  & --    & 55.0  & 51.0  & -& - & - & -\\
\midrule
InstructNav~\cite{long2024instructnav} & \checkmark & \checkmark & \checkmark & & 6.89 & --   & 31.0 & 24.0 & - & - & - & - \\

R2R-CMTP~\cite{chen2021topological}  & $\checkmark$ & $\checkmark$ & $\checkmark$ &  & 7.90  & 38.0  & 26.4  & 22.7  & - & - & - & - \\
LAW~\cite{raychaudhuri2021law}       &  & $\checkmark$ & $\checkmark$ & $\checkmark$ & 6.83  & 44.0  & 35.0  & 31.0  & 10.90 & 8.0 & 8.0 & 38.0 \\
CM2~\cite{georgakis2022cross}        &  & $\checkmark$ & $\checkmark$ & $\checkmark$ & 7.02  & 41.5  & 34.3  & 27.6  & - & - & - & - \\
WS-MGMap~\cite{chen2022weakly}       &  & $\checkmark$ & $\checkmark$ & $\checkmark$ & 6.28  & 47.6  & 38.9  & 34.3  & - & - & - & - \\
ETPNav + FF~\cite{wang2024sim}  &  & $\checkmark$ & $\checkmark$ & $\checkmark$ & 5.95  & 55.8  & 44.9  & 30.4  & 8.79 & 25.5 & 18.1 & - \\
Dynam3D \cite{wang2025dynam3d} &  & $\checkmark$ & $\checkmark$ & $\checkmark$ & 5.34 & 62.1 & 52.9 & 45.7 & - & - & - & - \\
NavFoM~\cite{zhang2025navfom} & $\checkmark$ & & & & 5.01 & 64.9 & 56.2 & 51.2 & 5.51 & 57.4 & 49.4 & 60.2 \\
OmniNav~\cite{xue2025omninav} & $\checkmark$ & & &  & \underline{3.74} & \underline{74.6} & \underline{69.5} & \textbf{66.1} & \textbf{3.77} & \textbf{73.6} & \underline{62.0} & - \\

\midrule

NaVid~\cite{zhang2024navid}   &  &  &  & $\checkmark$ & 5.47  & 49.1  & 37.4  & 35.9  & - & - & - & - \\

NaVILA~\cite{cheng2024navila}   &  &  &  & $\checkmark$ & 5.37  & 57.6  & 49.7  & 45.5  & - & - & - & - \\

UniNaVid~\cite{zhang2024uninavid}   &  &  &  & $\checkmark$ & 5.58  & 53.3  & 47.0  & 42.7  & 6.24 & 48.7 & 40.9 & - \\

StreamVLN~\cite{wei2025streamvln}   &  &  &  & $\checkmark$ & 4.98  & 64.2  & 56.9  & 51.9  & 6.22 & 52.9 & 46.0 & 61.9 \\
NavFoM \cite{zhang2025navfom} & & & &  $\checkmark$ & 5.01 & 64.9 & 56.2 & 51.2 & 5.51 & 57.4 & 49.4 & 60.2 \\
JanusVLN \cite{zeng2025janusvln} & & & & $\checkmark$ & 4.78 & 65.2 & 60.5 & 56.8 & 6.06 & 56.2 & 47.5 & 62.1 \\
DualVLN \cite{wei2025ground} & & & & $\checkmark$ & 4.05 & 70.7 & 64.3 & 58.5  & 4.58 &  61.4 & 51.8 & 70.0 \\
CorrectNav \cite{yu2025correctnav} & & & & $\checkmark$ & 4.24 & 67.5 & 65.1 & 62.3 & 4.09 & \underline{69.3} & \textbf{63.3} & \textbf{75.2} \\
\textbf{\methodname{} (Ours)} & & & & \checkmark & \textbf{3.64} & \textbf{76.4} & \textbf{69.9} & \underline{64.9} & \underline{3.90} & 68.0 & 59.3 & \underline{71.7} \\
\bottomrule
\end{tabular}
}
\label{tab:main_result}
\end{table*}

\subsection{Main Results}

\subsubsection{\textbf{Comparisons with state-of-the-arts on VLN-CE benchmarks}}
\begin{figure*}[h]
    \centering
    \vspace{+5pt}
    \includegraphics[width=\textwidth]{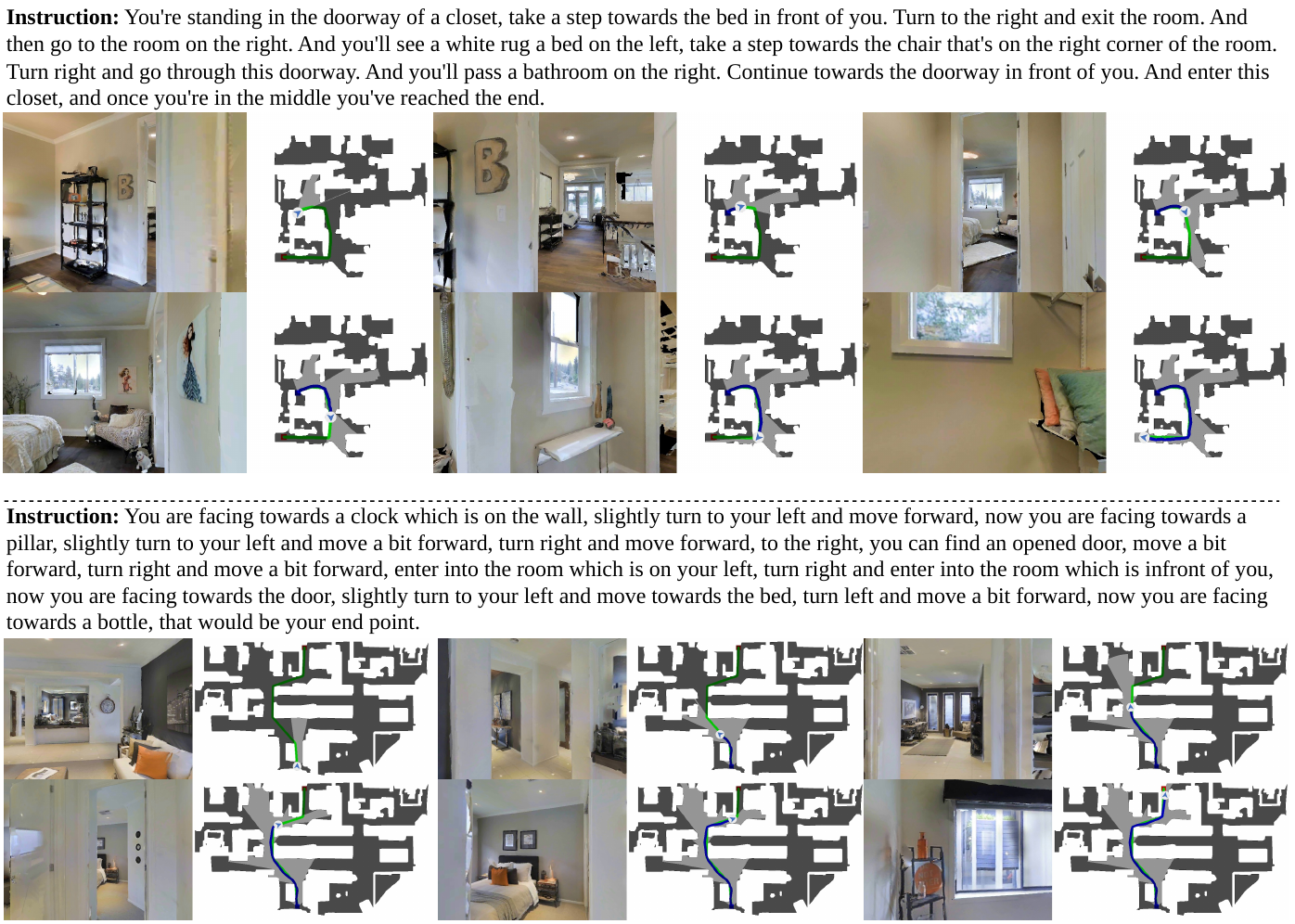}
    \vspace{-15pt}
    \caption{Qualitative results of JOP-VLN on the VLN-CE benchmarks.}
    \label{fig:simulation}
\end{figure*}

Table~\ref{tab:main_result} presents a comparison between JOP-VLN and state-of-the-art methods on the VLN-CE R2R and RxR Val-Unseen splits. Within the setting of using only single-view RGB observations, JOP-VLN achieves state-of-the-art performance on the R2R benchmark, outperforming the previous state-of-the-art method, CorrectNav~\cite{yu2025correctnav}, by 4.8\%, 8.9\%, and 2.6\% in terms of SR, OS, and SPL, respectively. Furthermore, JOP-VLN even outperforms methods employing panoramic views, such as OmniNav \cite{xue2025omninav} and NavFoM \cite{zhang2025navfom}.
In addition, our method obtains 68.0\% SR and 59.3\% SPL on RxR Val-Unseen split, demonstrating its excellent long-horizon decision-making capability.
Figure~\ref{fig:simulation} shows qualitative examples of JOP-VLN on the VLN-CE benchmarks.

\begin{figure}[!h]
    \centering
    \includegraphics[width=\columnwidth]{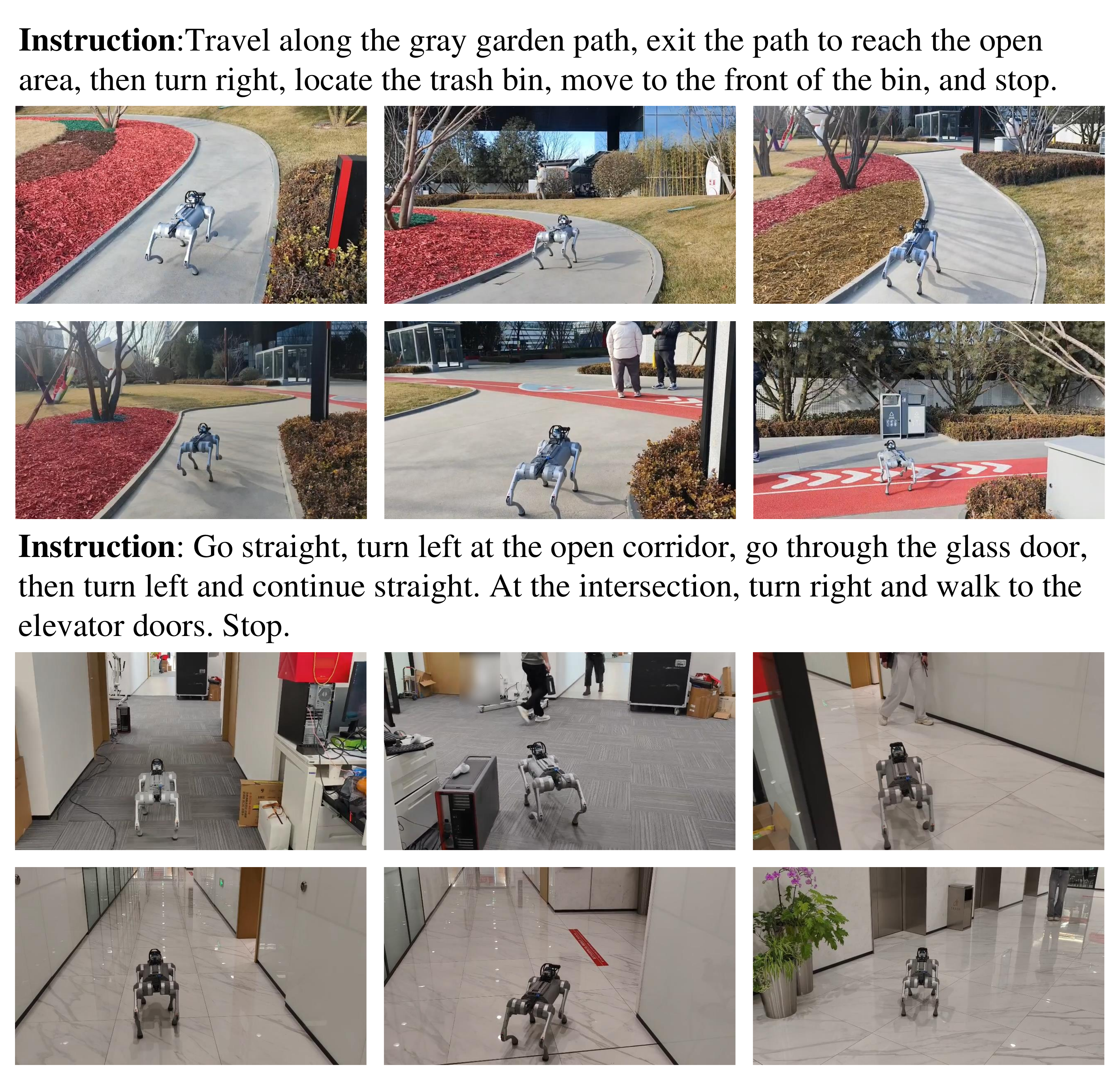}
    \vspace{-20pt}
    \caption{Qualitative results of JOP-VLN in representative real-world environments.}
    \label{fig:realworld}
\end{figure}

\subsubsection{\textbf{Real-world qualitative results}}
We evaluate the performance of JOP-VLN in physical environments with various characteristics, including an outdoor garden and an indoor office area. 
To bridge the significant domain shift between simulation and reality, we employ a few-shot adaptation strategy. Specifically, for each environment, we collect 10 navigation trajectories to perform 10-shot fine-tuning before testing. 
Qualitative examples of the model's trajectories are provided in Figure \ref{fig:realworld}. 
The results demonstrate that JOP-VLN consistently follows instructions accurately across both settings, underscoring its robustness and adaptability to real-world conditions.
Failure arises primarily from ambiguous referents in long instructions - when several visually similar landmarks (\textit{e.g.}, multiple doors along a corridor) co-exist, the agent occasionally enters the wrong one, leading to early termination or detours.

\subsection{Ablation Studies}
Table~\ref{tab:ablation} presents the ablation studies evaluating the effectiveness of our three-stage training pipeline. TS denotes trajectory summarization, while CHORD-ec refers to the use of the error-correction-prioritized trajectory sorting strategy for imitation learning within the CHORD framework.

\begin{table}[h]
\caption{Ablation Studies on VLN-CE R2R Val-Unseen split.}
\vspace{-20pt}
\label{tab:ablation}
\begin{center}
\begin{tabular}{c|ccc|cccc}
\toprule
\multirow{2}{*}{} & \multicolumn{3}{c|}{Training Strategy} & \multicolumn{4}{c}{R2R Val-Unseen}\\
 & Stage 1 & Stage 2 & Stage 3 & NE↓ & OS↑ & SR↑ & SPL↑ \\
\midrule
1 & IL & - & - & 5.90 & 59.2 & 49.5 & 43.6 \\
2 & IL+TS & - & - & 5.64 & 60.1 & 49.9 & 44.4 \\
3 & IL+TS & IL & - & 4.26 & 75.9 & 64.4 & 57.1 \\
4 & IL+TS & IL & IL & 3.99 & \textbf{76.8} & 68.1 & 60.7 \\
5 & IL+TS & IL & GRPO & 3.81 & 74.0 & 67.5 & 62.8 \\
6 & IL+TS & IL & CHORD & 3.84 & 75.0 & 68.7 & 63.7 \\
7 & IL+TS & IL & CHORD-ec & \textbf{3.64} & 76.4 & \textbf{69.9} & \textbf{64.9} \\
\bottomrule
\end{tabular}
\end{center}
\end{table}

\subsubsection{\textbf{Effectiveness of trajectory summarization (TS)}}
Comparing rows 1 and 2, the integration of trajectory summarization during the first stage leads to improved performance on the R2R benchmark. This demonstrates that navigational capabilities benefit from enhanced visual understanding.

\subsubsection{\textbf{Effectiveness of imitation learning on DAgger trajectories}}
Comparing rows 2 and 3, finetuning on the DAgger trajectories in the second stage yields a substantial improvement, with SR rising from 49.9\% to 64.4\%.
Furthermore, the integration of newly collected DAgger trajectories in Stage 3 further increases the SR to $68.1\%$. 
These results underscore the critical role of off-policy DAgger trajectories in mitigating covariate shift and enhancing error-recovery capabilities.

\subsubsection{\textbf{Effectiveness of CHORD with high-entropy sampling and error-correction-prioritised trajectory sorting (CHORD-ec)}}
In row 5, the application of reinforcement learning using the GRPO objective on high-entropy samples surpasses the post-training methods of pure imitation learning in terms of SPL and NE metrics, demonstrating the effectiveness of on-policy exploration.
In row 6, compared to pure RL in row 5, CHORD integrates the dual advantages of on-policy exploration and error recovery capabilities from off-policy expert demonstrations, outperforming both pure IL and pure RL methods.
Comparing row 7 to row 6, the inclusion of error-correction-prioritized trajectory sorting allows the model to more frequently encounter trajectories where it previously made mistakes, thereby enhancing its performance.

%% file: sec/5_conclusion.tex
\section{Conclusion}

In this paper, we introduce JOP-VLN, a novel vision-and-language navigation framework that synergistically integrates off-policy imitation learning and on-policy reinforcement learning. This joint learning approach harnesses the dual benefits of error recovery from expert demonstrations and trajectory exploration driven by reward signals. The high-entropy trajectory sampling strategy promotes exploration and prevents the model from entering a “zero-gradient” state caused by low-entropy samples, thereby enhancing the efficiency of RL training. Additionally, the error-correction-prioritized trajectory sorting strategy ensures that the model more frequently learns from trajectories where it previously made mistakes. Extensive experiments demonstrate the superiority of JOP-VLN, achieving new state-of-the-art performance on the VLN-CE R2R benchmark.